\documentclass[journal,onecolumn]{IEEEtran}

\usepackage{graphicx}
\graphicspath{{./images/}} 

\usepackage{url}
\usepackage{cite}
\usepackage{hyperref}
\usepackage{ amssymb }

\usepackage{mathtools}
\usepackage{times}
\usepackage{epsfig}
\usepackage{amsmath}
\usepackage{enumitem}
\usepackage[ruled]{algorithm2e}
\usepackage{array}
\usepackage{multirow}
\usepackage{color}
\usepackage{cases}
\usepackage[export]{adjustbox}
\usepackage[dvipsnames]{xcolor}
\usepackage{textpos}
\usepackage{comment}
\usepackage{textcomp}
\usepackage{listings}
\usepackage{bibentry}
\usepackage{booktabs}

\usepackage{rotating}

\usepackage{comment}
\usepackage[dvipsnames]{xcolor}

\hypersetup{colorlinks=true}

\begin{document}
%
\title{Event Detection in Football using Graph Convolutional Networks}
%
%
%

\author{ Aditya Sangram Singh Rana
\thanks{Author: Aditya Sangram Singh Rana, adityasangramsingh.rana@e-campus.uab.cat}
\thanks{Advisor 1: Dr. Francesc Moreno Noguer,  Institut de Robòtica i Informàtica Industrial,  Universitat Politècnica de Catalunya }
\thanks{Advisor 2: Dr. Antonio Rubio Romano,  Computer Vision and Machine Learning Team,  Kognia Sports Intelligence }
\thanks{Thesis dissertation submitted: September 2021}}

\markboth{Master Thesis Dissertation, Master in Computer Vision, September 2021}%
{Shell \MakeLowercase{\textit{et al.}}: Bare Demo of IEEEtran.cls for Journals}

\maketitle

\begin{abstract}
The massive growth of data collection in sports has opened numerous avenues for professional teams and media houses to gain insights from this data. The data collected includes per frame player and ball trajectories, and event annotations such as passes, fouls, cards, goals, etc. Graph Convolutional Networks (GCNs) have recently been employed to process this highly unstructured tracking data which can be otherwise difficult to model because of lack of clarity on how to order players in a sequence and how to handle missing objects of interest. In this thesis, we focus on the goal of automatic event detection from football videos. We show how to model the players and the ball in each frame of the video sequence as a graph, and present the results for graph convolutional layers and pooling methods that can be used to model the temporal context present around each action.
\end{abstract}


\IEEEpeerreviewmaketitle

\section{Introduction}
\label{section:introduction}

\begin{figure}[t]
    \centering
    \includegraphics[width=\linewidth]{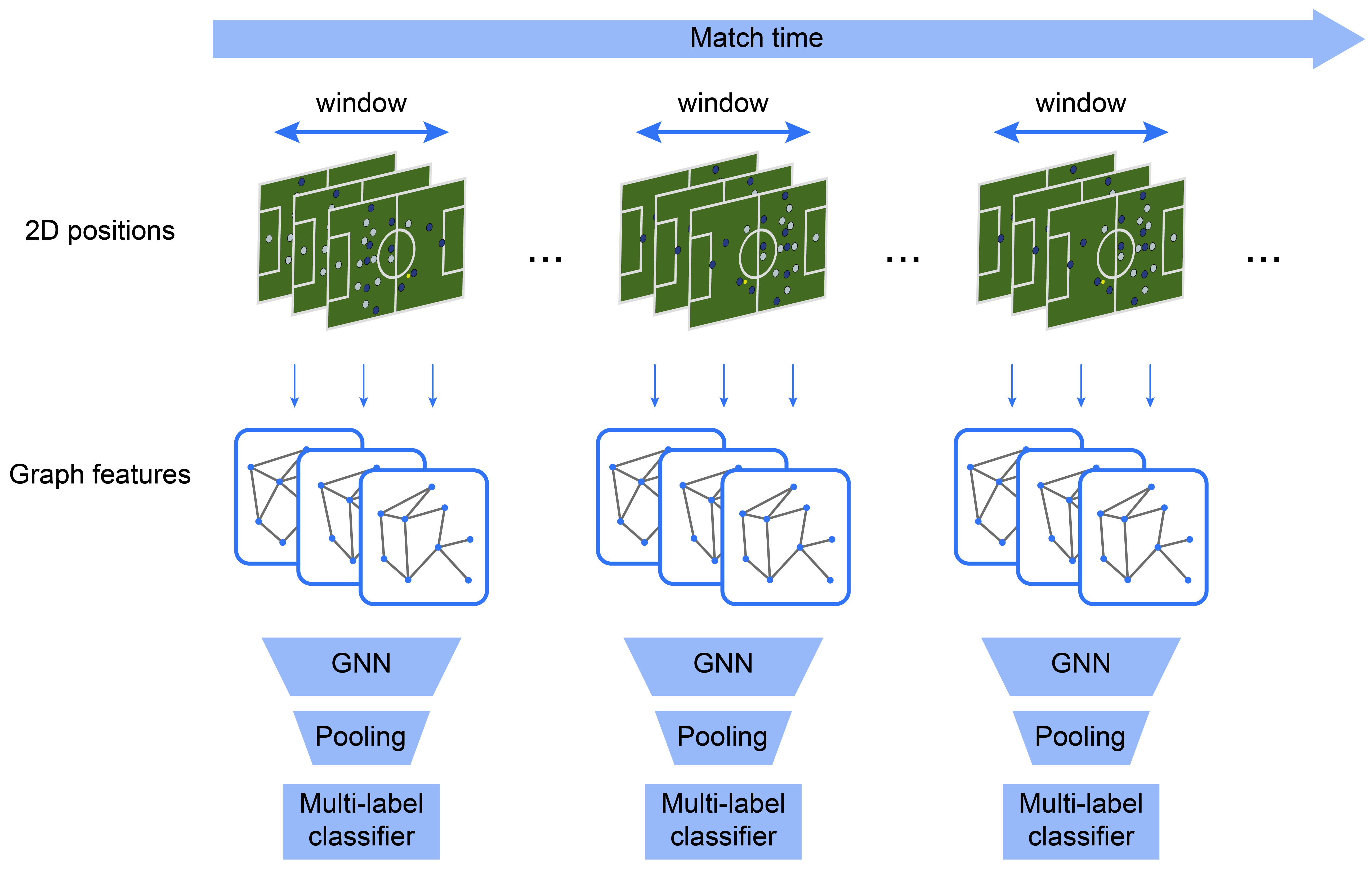}
    \caption{
    \textbf{Overview of the proposed method.} 2D positions of players and ball in each frame are represented as graphs, that are then fed into a Graph Neural Network architecture to extract features. These features are then pooled taking into account temporal windows and the pooled features are fed into a multi-label classifier.
    }
    \label{fig:summary}
\end{figure}

\IEEEPARstart{S}{ports} have emerged to be the highest revenue-generating applications of computer vision \cite{StatistaCV} with an annual market revenue of \textdollar 91 billion \cite{91billion}, with \textdollar 28.7 billion \cite{28billion} generated solely from the European football industry, mostly from broadcasting and commercial activities. It has played a big role in enhancing the visual experience of live sports broadcasting through the integration of augmented and virtual reality. Machine learning has also revolutionized the sports industry in the way athletes train, how their game performances are analyzed and how coaches prepare their teams to tackle an opposition. For both sports analysts and broadcast producers, it is crucial to be able to identify and summarize all the events that occur within a game. Currently, this requires hours of manual annotation and a high-level understanding of the game being played. With around 10,000 football games scheduled for the five biggest leagues in Europe every year, creating an automatic event detection system would save hundreds of thousands of hours of manual annotation and speed up the process by a huge factor. It would also help in cutting down the high costs of production that can only be afforded by the top leagues and that in turn leaves behind a majority of the games from lower leagues and less popular sports uncovered.

\begin{figure}[t]
\begin{center}
\includegraphics[width=.8\textwidth]{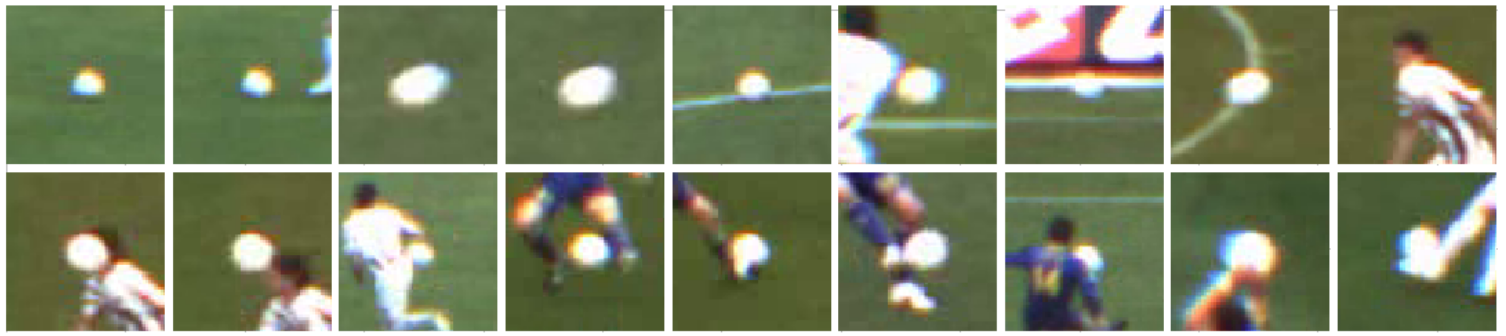}
\end{center}
\caption{Example of patches illustrating high variance in ball appearance and difficulty of the ball detection task. Image taken from \cite{DeepBall}.}
    \label{fig:ball_difficulty}
\end{figure}

One of the key elements in automating any sports' statistical analysis is an accurate and efficient ball-player detection system. Detecting the ball from the long-shot scene of a football game is a challenging task. Komorowski et al., 2019 \cite{DeepBall} discusses several factors that make localizing the ball on the football pitch difficult. The ball has a very small size compared to all the other objects present on the pitch and in the scene. Also, its size can vary a lot depending on its position on the pitch. The size of the ball can be as small as $8 \times 8$ pixels in a $1980 \times 1080$ or a $3840 \times 2160$ pixel image depending on the camera resolution. This can force the detector to output lots of false-positive values for the ball as it may be difficult to differentiate the ball from the white socks and white shoes of the players, sometimes even being mistaken for a bald person's head \cite{baldreferee}. Also, when in motion the ball may appear as blurry and elliptical instead of its original circular shape as can be seen in Fig. \ref{fig:ball_difficulty}. Detecting the players on the pitch is easier compared to the ball since they are much larger in size. However, it may be difficult to detect players when they are occluded by another player. 

\begin{figure}[t]
\begin{center}
\includegraphics[width=0.9\textwidth]{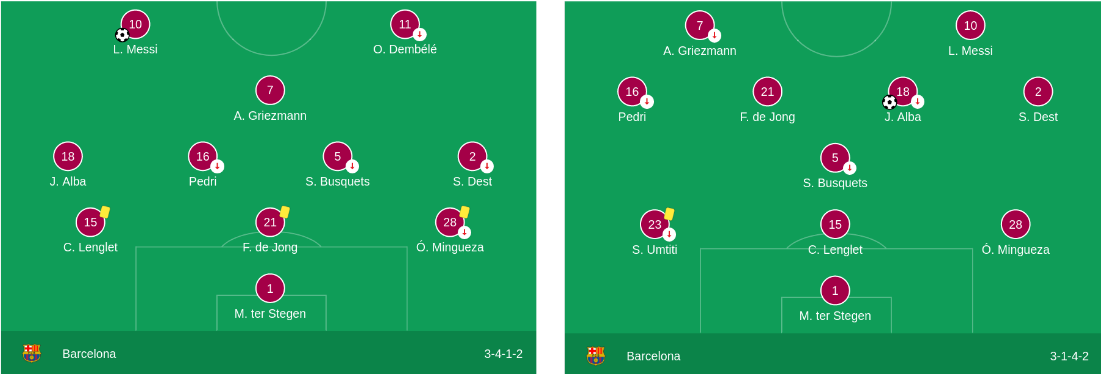}
\end{center}
\caption{Different teams can use different formations in the same season, as can be seen above for a team like FC Barcelona that has recently been changing its style and formation of play quite regularly. Modelling the motion of players of a team becomes difficult when you use models that predefine a specific formation template \cite{lucey_large_scale, lucey_not_all_passes_equal} when trying to analyze a team or generating features for another downstream task.}
    \label{fig:formation}
\end{figure}

Positional data from football games is difficult to model because most machine learning algorithms require data and features to be arranged in a specific order. It is unclear how to construct a feature vector taking into account the individual features of the players and the ball as there are no clearly defined rules on what order to use for selecting the players. Lucey et al. tried to overcome this ordering issue by predefining a specific formation template of 4-3-3 to each team \cite{lucey_large_scale, lucey_not_all_passes_equal} and then assigning each player a role in this arrangement. However, this model can easily fail its assumptions as teams use a variety of formations, so the roles of the defenders and attackers playing in a 3-5-2 formation would be very different from a 4-3-3 formation. Hence, we would not be modeling the roles of the players correctly as can be seen in Fig. \ref{fig:formation}. Also, this formation template requires all players to be present on the pitch, and cannot be used when a player from a team has been sent off. 

Other papers \cite{prob_movement_model, soccermap} have tried working with image representations of tracking data to overcome this ordering/alignment and then processing them using Convolutional Neural Networks (CNNs). However, taking the tracking data which is a compact, complete and low-dimensional representation of an object's movement and converting it to an image that is a sparse high-dimensional representation is sub-optimal. Modeling our data with Graph Neural Networks (GNNs) can help us overcome these issues as
\begin{itemize}
    \item we can neglect the need for ordering the features in a certain order
    \item deal with a variable number of players on the pitch, and can also handle missing tracks for the ball or players
    \item learn local and high-level features directly from the tracking data
\end{itemize}

GNNs have recently gained a lot of popularity and have been employed in a diverse set of domains where data is more aptly represented as graphs or networks. For example, in chemistry molecules are modeled as graphs with atoms as nodes and the bonds representing the edges ~\cite{duvenaud2015convolutional, gilmer2017neural}. In a social network, the interactions between the users can help us determine which accounts are fake or bots~\cite{kipf2017semi,monti2019fake}. Bronstein et al \cite{geometricdeeplearningbook} discuss in detail the history and development of the field of machine learning on graphs in their book on geometric deep learning \cite{geometricdeeplearningbook}. With the rapid development of ideas in this field, it is important to benchmark the performance of existing architectures against large datasets under consistent experimental settings, a task that has been recently achieved in the paper \cite{benchmarking_gnns} by Dwivedi et al.

In the following sections, we will describe the generation process and the structure of our input data, from frames obtained from the camera to the real-world coordinates of all the entities (referring to football players, referees and the ball) present on the football pitch. However, the main focus of this thesis is the processing of this tracking data. The contributions of this thesis can be summarized as follows
\begin{enumerate}
    \item Formulating the pipeline for generating a high-level view of the football field (referred to as a minimap) using ball-player bounding boxes and camera calibration
    \item Describing how football tracking data can be modelled using graphs and then processed using Graph Convolutional Networks
    \item  Formulating event detection as an action spotting task, which involves localizing events to a certain timestamp in a video
    \item Experimenting with different pooling methods for modelling the temporal context around each action
\end{enumerate}
\section{State of the art}

\subsection{Object Detection} Komorowski et al. \cite{footandball} compare the performance of state-of-the-art object detectors on publicly available football datasets ISSIA-CNR Soccer \cite{issia_cnr} and Soccer Player Detection \cite{soccerdetection}. They also propose their own architecture, inspired by the Faster-RCNN (Region-based Convolutional Neural Networks) \cite{faster_rcnn} and the Feature Pyramid Network (FPN) \cite{fpn} proposed by Girsick et al., that can run at a much faster framerate while being at par on performance with these networks. Their network, called FootandBall \cite{footandball}, has 200K parameters compared to the 41M parameters in FPN Faster-RCNN with a Resnet-50 backbone \cite{resnet}, and runs at rate of 37 fps (vs 8fps of the latter) when processing a high-definition ($1920 \times 1080$) video. There is a need for an updated study that compares the performance of the new and improved single-stage architectures \cite{retina, yolov4} and transformer-based architectures \cite{detr, swin} on these datasets. The trade-off between single-stage and two-stage architectures is that of speed vs accuracy. Two-stage detectors, like Faster-RCNN split the detection task into region proposal and then regression-classification. One-stage detectors, on the other hand, perform bounding box regression and classification directly on the image. Two-stage architectures usually perform better on smaller objects as they get a chance to look at the image twice, and refine the proposals from the first-stage instead of having to regress the coordinates in a single pass.

\subsection{Action Spotting}
Certain actions like shot on goal, pass, offside do not happen over a time window but can be anchored to a certain frame/time that defines the event. For example, in soccer, a goal happens at the exact moment the football crosses the goal line. The closer our predictions are to the target, the better our action-spotting performance is. The task of action spotting was introduced by Giancola et al. in their SoccerNet paper \cite{soccern_net} along with the SoccerNet dataset in Computer Vision and Pattern Recognition (CVPR) conference 2018. The SoccerNet dataset consists of 500 football games (amounting to 764 hours of video footage) from the main European leagues collected across three seasons. They provided event annotations for the games by splitting them into three main categories: 'goals', 'cards' and 'substitutions', for a total of 6,637 events. In the year 2021, they released the second version of this dataset dubbed SoccerNet v2.0 \cite{soccern_net_v2}, where they increased the number of actions annotated to 17. They introduced new classes corner, throw-in, shots on target, offside, etc. and increased the number of available annotations to 110,458, with an average of 221 actions per game. Current state of the art methods for action spotting include Context Aware Loss Function (CALF) \cite{calf_loss, calf_gcn} and NetVLAD++ \cite{netvlad_plus_plus}, the latter of which we are going to be using for our experiments.

\subsection{Graph Convolutional Networks}

\begin{figure}[t]
\begin{center}
\includegraphics[width=0.9\textwidth]{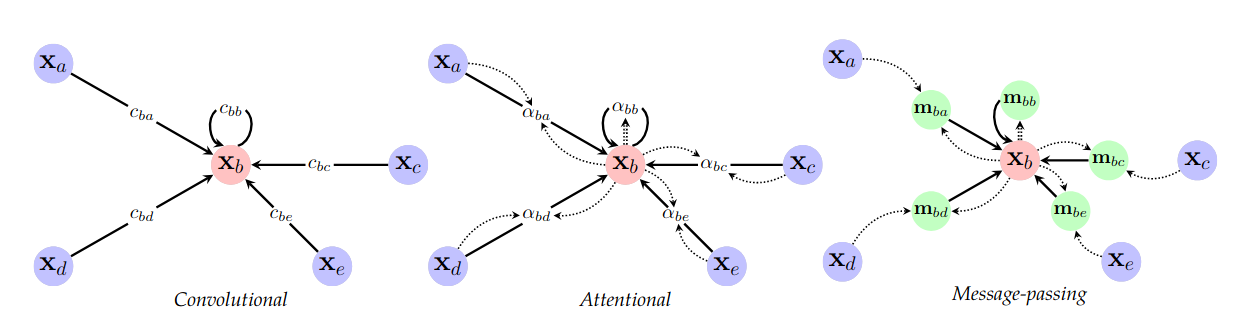}
\end{center}
\caption{A visualisation of the dataflow for the three flavours of GNN layers, g. We use the neighbourhood of node b from Figure 10 to illustrate this. Left-to-right: convolutional, where sender node features are multiplied with a constant, $c_{uv}$; attentional, where this multiplier is implicitly computed via an attention mechanism of the receiver over the sender: $\alpha_{uv} = a(x_u,x_v)$; and message-passing, where vector-based messages are computed based on both the sender and receiver: $m_{uv} = \psi(x_u,x_v)$. Retrieved from \cite{geometricdeeplearningbook}}
    \label{fig:gcn_flavour}
\end{figure}

GNNs are amongst the most general class of deep learning architectures as most other learning architectures can be formulated as a special case of GNNs with additional geometric structures. A class of GNNs called Graph Convolutional Networks (GCNs) generalize the idea of convolution from euclidean to the graph domain. The same GCN model has been developed parallelly between different streams of literature, the popular dichotomy being the one between spectral theory on graphs \cite{chung_book, bruna2013spectral, NIPS2016_6081, kipf2017semi,levie2019cayleynets} and spatial-based graph convolutions. Regardless of the motivation, the defining feature of a GCN is that it uses a form of neural message passing in which vector messages are exchanged between nodes and updates using neural networks \cite{gilmer2017neural}. Different flavours of this model can be seen in Fig. \ref{fig:gcn_flavour}.

Recent works \cite{benchmarking_gnns} on benchmarking the performance of GNNs have shown that anisotropic mechanisms \cite{battaglia2016interaction,marcheggiani2017encoding,Monti_2017,velickovic2018graph,bresson2017residual} improve the performance of GCNs. The best results amongst this class of models were achieved by GAT \cite{velickovic2018graph} which leverages attention \cite{attention_vaswani} and GatedGCN \cite{bresson2017residual} that uses gating introduced in Chung et al \cite{gated_gru}.

One of the key reasons behind the success of Convolutional Neural Networks (CNNs) \cite{lecun_cnn, lecun_gradient} is the design and training of very deep models by stacking many layers together with residual connections \cite{resnet} between them. Stacking more than four layers in a vanilla GCN leads to an over-smoothing problem in which deeper node features converge to the same value and local neighborhood information is lost. Li et al. \cite{deepgcns} mitigate this problem by adapting the ideas of ResNets \cite{resnet} and DenseNets \cite{densenet} to construct deep GCNs up to fifty-six layers achieving state of the art on semantic segmentation of point clouds.

\section{Method} \label{sec:method}

In this section, we describe how we encode per-frame player and ball tracking data in a graph, and investigate several temporal pooling methods that learn the past and future context independently to perform the task of action spotting.

\subsection{Graphs}

\begin{figure}[t]
\begin{center}
\includegraphics[width=.8\textwidth]{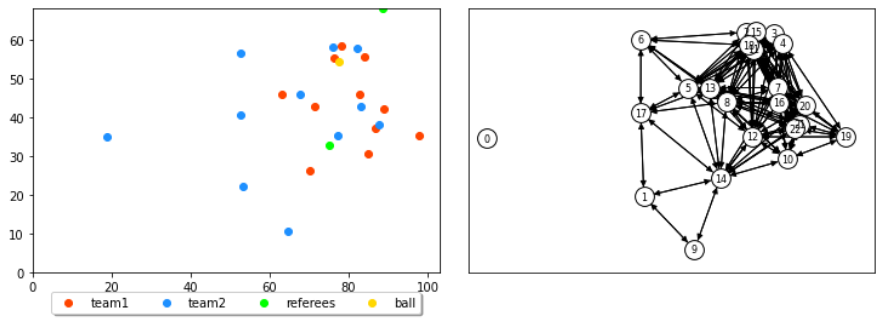}
\end{center}
\caption{\textbf{Ball-Player Graph} In our graph, the ball and the players are represented by nodes, connected through an edge if the real-world distance between them is less than 25 meters, which we consider sufficient for message passing. The feature for each node is constructed by concatenating its $x$ and $y$ position (normalized to $[-0.5, 0.5]$) based on the pitch size of the field, and a one-hot vector representing if the node is a player from the first team, second team or the ball. The choice of the first team and second team is irrelevant as the important idea is that players from the same team have the same one-hot vector label. At this point, I would like to reiterate that only about 12.5 percent of each game in our data has ball annotations and for the rest of the cases, the model has to learn and make predictions solely based on the movements of the players.}
    \label{fig:graph}
\end{figure}

A graph is a ubiquitous data structure and can be described as a collection of objects that may or may not interact with each other. Formally, a graph $\mathcal{G} = (\mathcal{V}, \mathcal{E})$ is defined by a set of nodes $\mathcal{V}$ connected by a set of edges $\mathcal{E}$. We denote an edge going from node $u \in \mathcal{V}$ to node $v \in \mathcal{V}$ as $(u,v) \in \mathcal{E}$. A convenient way to represent graphs and its node features is through an adjacency matrix $A \in {R}^{ |\mathcal{V}| \times |\mathcal{V}|}$ and a feature matrix $F\in {R}^{ |\mathcal{V}| \times m}$ respectively, where $|\mathcal{V}|$ is the number of nodes/vertices in the graph and $m$ is the dimensionality of the node feature vector. When constructing the feature matrix $F$, we assume that the ordering of the nodes is consistent with the ordering in the adjacency matrix $A$. Even though in most graphs $A$ is usually a binary matrix of $\{0,1\}$ signifying whether an edge exists between two nodes or not, it can also be a real matrix or a matrix of vectors $A \in {R}^{|\mathcal{V}| \times E \times |\mathcal{V}|}$, where each edge can be weighted or have its own vector representation of length $E$. The way we construct the graphs for our models is depicted in Fig. \ref{fig:graph}.

\subsection{NetVLAD}

\textbf{VLAD} (Vector of Locally Aggregated Descriptors), 2010 \cite{vlad} was presented as a solution to large-scale image retrieval which was restricted by memory requirements at the time, as an alternate to the Bag of Words (BoW) \cite{bow} and Fisher Vectors \cite{fisher} approach. VLAD, like visual word encoding, starts by vector quantizing feature descriptors. It differs from BoW as instead of keeping the count of visual words, VLAD stores the sum of residuals (difference vector between the descriptor and its corresponding cluster center). Mathematically, given a set of $N$ $D$-dimensional features $\{\mathbf{x}_i\}_{i=1..N}$ as input, a set of $K$ clusters centers $\{\mathbf{c}_k\}_{k=1..K}$ with same dimension $D$ as VLAD parameters, the output of the VLAD descriptor $V$ is defined by:

\begin{equation}
    V(j,k) = \sum_{i=1}^N a_k(\mathbf{x}_i)(\mathbf{x}_i(j)-\mathbf{c}_k(j))
    \label{eq:vlad}
\end{equation}

where $\mathbf{x}_i(j)$ and $\mathbf{c}_k(j)$ are respectively the $j$-th dimensions of the $i$-th descriptor and $k$-th cluster center. $a_k(\mathbf{x}_i)$ denotes the hard assignment of the sample $\mathbf{x}_i$ from its closer center, i.e. $a_k(\mathbf{x}_i)=1$ if $\mathbf{c}_k$ is the closest center of $\mathbf{x}_i$, $0$ otherwise. The matrix $V$ is then L2-normalized at the cluster level, flatten into a vector of length $D \times K$ and further L2-normalized globally.

\textbf{NetVLAD.} The VLAD module is non-differentiable because of the hard assignment $a_k(\mathbf{x}_i)$ of the samples $\{\mathbf{x}_i\}_{i=1}^N$ to the clusters $\{\mathbf{c}_k\}_{i=1}^K$. These hard assignments create discontinuities in the feature space between the clusters, hindering us from learning the parameters of VLAD by backpropagation. To overcome this issue, NetVLAD introduced a soft-assignment $\tilde{a}_k(\mathbf{x}_i)$ for the samples $\{\mathbf{x}_i\}_{i=1}^N$, based on their distances from each cluster center. Formally:

\begin{equation}
    \tilde{a}_k(\mathbf{x}_i) = \frac { e^{-\alpha \| \mathbf{x}_i - \mathbf{c}_k\|^2} } { \sum_{k'=1}^K e^{-\alpha \| \mathbf{x}_i - \mathbf{c}_{k'}\|^2} }
    \label{eq:softassignment}
\end{equation}

$\tilde{a}_k(\mathbf{x}_i)$ ranges between $0$ and $1$, with the highest value assigned to the closest center. $\alpha$ is a temperature parameter that controls the softness of the assignment, a high value for $\alpha$ (ex. $\alpha \rightarrow + \infty$) would lead to a hard assignment like in VLAD. Furthermore, by expanding the squares and noticing that $ e^{-\alpha \| \mathbf{x}_i \|^2} $ will cancel between the numerator and the denominator, we can interpret Equation~\ref{eq:softassignment} as the softmax of a convolutional layer for the input features parameterized by $\mathbf{w}_k = 2 \alpha \mathbf{c}_k$ and $b_k = - \alpha \| \mathbf{c}_k \| ^2$. Formally:

\begin{equation}
    \tilde{a}_k(\mathbf{x}_i) = \frac { e^{\mathbf{w}_k^T \mathbf{x}_i + b_k} } { \sum _{k'} e^{\mathbf{w}_{k'}^T \mathbf{x}_i + b_{k'}} }
    \label{eq:convsoftmax}
\end{equation}

Finally, by plugging the soft-assignment from \ref{eq:convsoftmax} into the VLAD formulation in \ref{eq:vlad}, the NetVLAD features are defined
as in Equation \ref{eq:netvlad}, later L2-normalized per cluster, flattened and further L2-normalized in its entirety.

\begin{equation}
    V(j,k) = \sum_{i=1}^N\frac { e^{\mathbf{w}_k^T \mathbf{x}_i + b_k} } { \sum _{k'} e^{\mathbf{w}_{k'}^T \mathbf{x}_i + b_{k'}} } (\mathbf{x}_i(j)-\mathbf{c}_k(j))
    \label{eq:netvlad}
\end{equation}

\begin{figure}[t]
\begin{center}
\includegraphics[width=.9\textwidth]{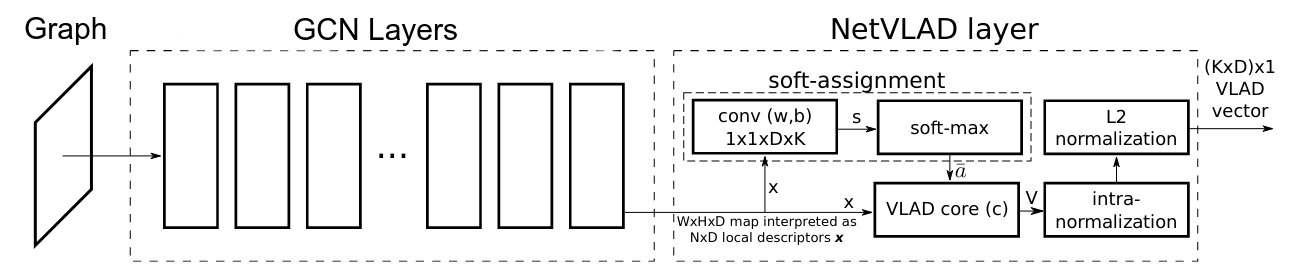}
\end{center}
\caption{\textbf{Graph architecture}  with the NetVLAD layer. The layer can be implemented using standard CNN layers (convolutions, softmax, L2-normalization) and one easy-to-implement aggregation layer to perform aggregation in equation \ref{eq:vlad}, joined up in a directed acyclic graph. Parameters are shown in brackets.}
\label{fig:netvlad_architecture}
\end{figure}

While the original VLAD optimizes solely the cluster centers $\mathbf{c}_k$, NetVLAD optimizes for three sets of independent parameters $\{\mathbf{w}_k\}$, $\{b_k\}$ and $\{\mathbf{c}_k\}$, dropping the constraint of $\mathbf{w}_k = 2 \alpha \mathbf{c}_l$ and $b_k = - \alpha \| \mathbf{c}_k \| ^2$. All parameters of NetVLAD are learnt for the specific task in an end-to-end manner. As illustrated in Fig. \ref{fig:netvlad_architecture}, the NetVLAD layer can be visualized as a meta-layer that is further decomposed into basic CNN layers connected up in a directed acyclic graph, and can be easily plugged into any architecture for training. We also test our approach using NetRVLAD \cite{soccern_net, netvlad_plus_plus} which is a slightly tweaked version developed on top of NetVLAD, which drops the cluster parameters $\mathbf{c}_k(j)$ in \eqref{eq:netvlad}, leading to slightly less parameters to learn.
 
\subsection{Pooling for Detection}

In order to recognize or detect activities within a video, a common practice consists of \textbf{aggregating} local features and \textbf{pooling} them. While naive approaches use mean or maximum pooling, more elaborate techniques such as Bag-of-Words (BOW)~\cite{bow}, Fisher Vector (FV)~\cite{fisher}, and VLAD~\cite{vlad} look for a structure in a set of features by clustering and learning to pool them in a manner that improves discrimination. Recent works extend those pooling techniques by incorporating them into Deep Neural Network (DNN) architectures, namely SoftDBOW~\cite{philbin_lost}, NetVLAD~\cite{netvlad,actionvlad}, NetRVLAD\cite{netvlad_plus_plus} and ActionVLAD~\cite{actionvlad}.

We follow an action spotting pipeline similar to the one proposed in SoccerNet \cite{soccern_net}, where they try to classify if an action lies within a certain time window in a multi-label setting. During training, we split our videos into chunks of different lengths annotated with all events occurring within that time window. We aggregate the features extracted from all frames present within that window and use a sigmoid activation at the last layer, as is usual in the task of multi-label classification. For testing, we sample the frames of the video with the same window size and a stride of one and pass this as input to the model to obtain the raw event predictions for each frame. We then apply a confidence threshold and non-maximum suppression on the output predictions for each class to get the final action spotting results.

\begin{figure}[t]
\begin{center}
\includegraphics[width=0.5\textwidth]{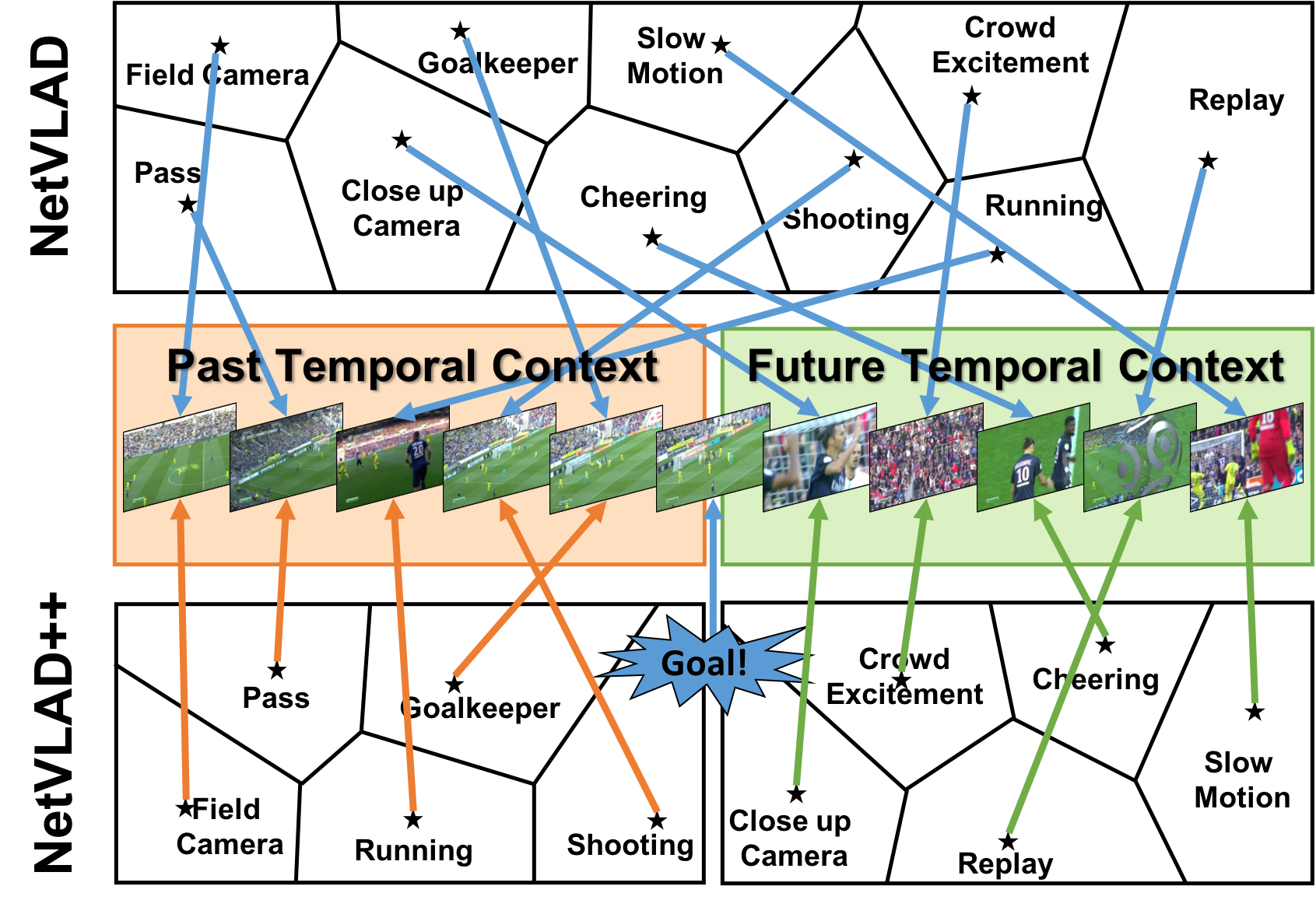}
\end{center}
\caption{NetVLAD (top) vs. NetVLAD++ (bottom) pooling
modules for action spotting. The temporally-aware NetVLAD++
pooling module learns specific vocabularies for the past and future
semantics around the action to spot. Figure taken from \cite{netvlad_plus_plus}.}
\label{fig:netvlad}
\end{figure}

\subsection{Temporal Pooling - Layer ++}
\label{subsection:netvlad++}

All the above-mentioned pooling methods are permutation invariant and do not take into account the order of the frames, hence losing the temporal information. Recent works \cite{calf_loss,calf_gcn, netvlad_plus_plus} have shown that temporal context before and after an event is very different and should be handled differently. They describe how different actions might share the same similar sets of features before or after the event but usually not both. For example, the semantic information before a 'goal' event and a 'shot on goal' are very similar, representing the concept of a player trying to score a goal and a goalkeeper trying to catch the ball. Yet, the semantic information derived from the movement of the players after the two events is very different, as the goal is usually followed by all the players gathering together. For our experiments, we use the idea proposed in \cite{netvlad_plus_plus}, where we add two separate pooling modules for aggregating features from before and after the action separately, as can be seen in Fig. \ref{fig:netvlad}. The comparison of performance between the pooling layers and the temporally aware pooling layers is provided in the table \ref{tab:results} where ++ represents that the layer is used in temporal pooling fashion. 

We define the
 \textit{past} context as the frame feature with a temporal offset in $[-T_b, 0]$ and the \textit{future} context as the frame feature with a
 temporal offset in $[0, T_a]$. Each pooling module aggregates different clusters of information from the $2$ subsets of features, using $K_a$
 and $K_b$ clusters, respectively for the \textit{after} and \textit{before} subsets. Formally:

\begin{equation}
\begin{split}
    V = AGGREGATE( V_b, V_a )
\end{split}
    \label{eq:netvlad++}
\end{equation}

where $AGGREGATE$ is an aggregation functio $V_b$ and $V_a$ that represents the NetVLAD pooled features for the sample \textit{before} and \textit{after}
the action occurs, parameterized with $K_b$ clusters for the \textit{past} context and $K_a$ clusters for the \textit{future} context.

\section{Data Generation}\label{sec:data}
One of the main motivations for this thesis is being able to leverage information from a large set of football matches without the burden of processing each frame of the videos. The results shown in this paper are based on simple 2D tracking data for the players and the ball. But, of course, some preprocessing work is required to obtain this low-dimensional data. This section describes the data used in section~\ref{sec:method}, as well as the pipeline used to get it from the initial football videos.

\begin{figure}[t]
    \centering
    \includegraphics[width=0.95\linewidth]{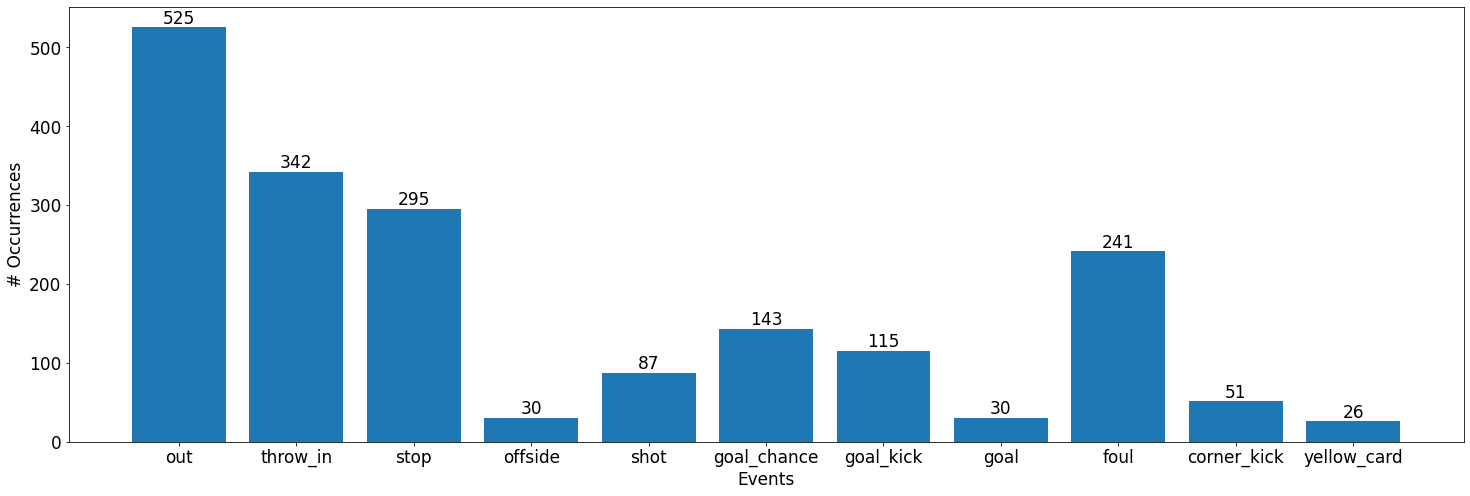}
    \caption{ \textbf{Dataset distribution.} The bar plot shows the number of annotations available for each event in the dataset. Though we do not have any annotations available for a red\_card in our dataset yet, we have included it in our model because it will be be annotated eventually as more games are recorded and become available for training.}
    \label{fig:dataset_distribution}
\end{figure}

\subsection{Preprocessing}
\begin{figure}[t]
    \centering
    \includegraphics[width=\linewidth]{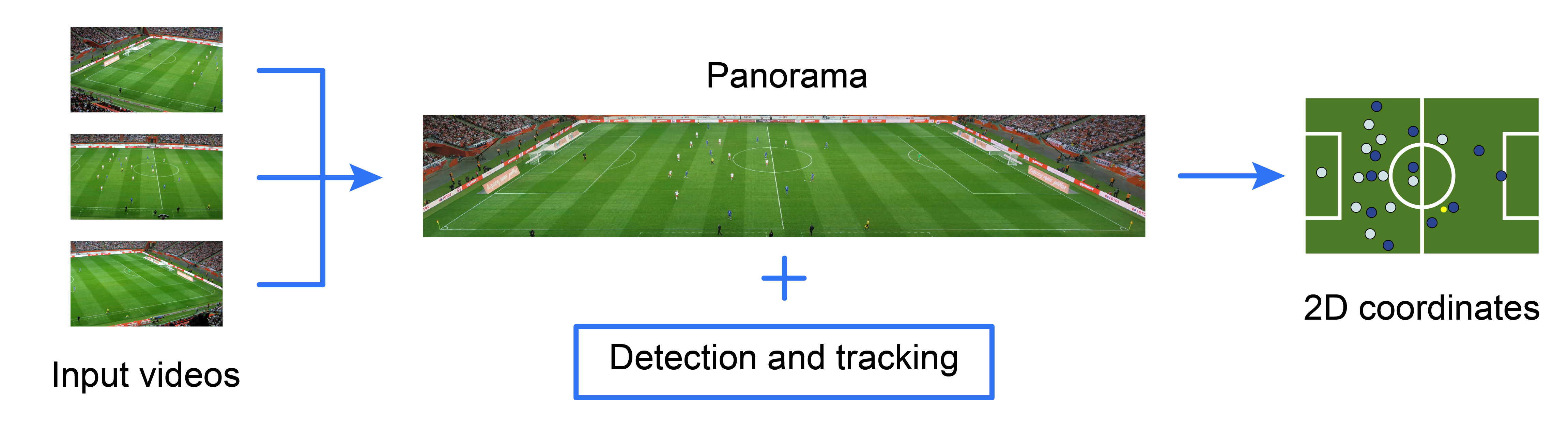}
    \caption{
    \textbf{Summary of the data preprocessing pipeline.} Videos coming from three cameras are joined to form a \textit{panorama}, which is calibrated with respect to a map of the pitch, where we project the detected players and ball position to get a compact representation of the information in each frame.
    }
    \label{fig:data_pipeline}
\end{figure}

Steps to convert video data (3 images with 4K resolution per each frame of the match at a frame rate of 15fps) into low-dimensional 2D tracking data (23 $xy$ positions for players and the ball per frame, with team information for each player) are detailed below (see Fig.~\ref{fig:data_pipeline} for a visual summary):

\begin{enumerate}
    \item For any specific game, the starting point are three videos covering the left, central and right parts of the pitch with some overlap between them.
    \item With these videos, a bigger one where the whole pitch is visible (referred as the \textit{panorama video} from now on) is constructed after a calibration process.
    \item A Faster-RCNN~\cite{faster_rcnn} architecture with a ResNet~\cite{resnet} backbone trained with football images is used to detect the players and the ball in the video.  
    \item Features for re-identification extracted based on~\cite{zhou2019omni}.
    \item Hungarian algorithm~\cite{kuhn1955hungarian} to associate detections with trajectories based on their features, their position in the previous frame and and an estimation of their current position using a Kalman filter~\cite{rudolf1960kalman}.
    \item Players, referees and ball positions are projected onto the football pitch minimap.
\end{enumerate}

\subsection{Information in the dataset}
Our data includes information from 9 matches recorded at 15 fps with the following information:
\begin{itemize}
    \item Names and IDs of the teams and players of the match.
    \item Name of the stadium and size of the pitch.
    \item 2D position of each player and referee on the pitch in meters for each frame of the match.
    \item 2D ball position on the pitch in meters for 10-12\% of the frames of the match (without 3D information, so there's a reprojection error when the ball is not touching the ground).
    \item Frame number where certain events happen, associated with the corresponding player ID when needed. Concretely, these are the annotated events and their distribution can be found in \ref{fig:dataset_distribution}.
    \begin{enumerate}
    \item game flow annotations:
        These are used to note when the game stops. These tell the time of the play being stopped
        but interpretation will depend on what annotation follows (Except goal, the interpretation
        is clear for goal).
        \begin{itemize}
        \item 01. out
        \item 02. stop
        \item 03. goal
        \end{itemize}
    \item post-out annotations:
        These always come between "out" and "pass". The exact time of these doesn't matter; they
        are just used to tell us how to interpret the next "pass".
        \begin{itemize}
        \item 04. goal\_kick
        \item 05. corner\_kick
        \item 06. throw\_in
        \end{itemize}
    \item post-stop annotations:
        These work similarly as out events, but follow "stop" rather than "out". These can be
        followed by "pass" or "shot".
        \begin{itemize}
        \item 07. offside
        \item 08. foul
        \item 09. yellow\_card
        \item 10. red\_card
        \end{itemize}
    \item other:
        \begin{itemize}
        \item 11. goal\_chance - when the previous annotation (touch event) was a chance at goal
        \item 12. shot
        \end{itemize}
    \end{enumerate}
\end{itemize}

\begin{figure}[t]
\begin{center}
\includegraphics[width=.4\textwidth]{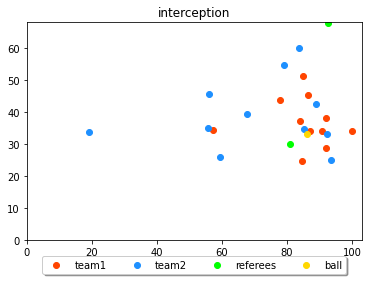}
\includegraphics[width=.4\textwidth]{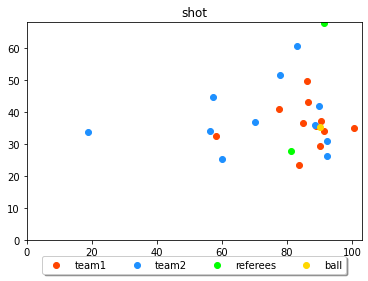}
\includegraphics[width=.4\textwidth]{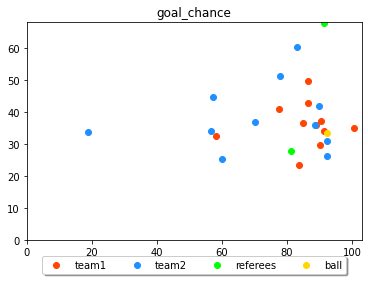}
\includegraphics[width=.4\textwidth]{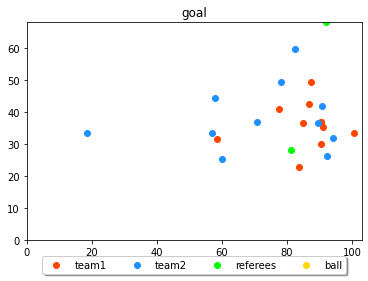}
\end{center}
\caption{The above frames give an example of the way frames are annotated right before a goal.}
\label{fig:events}
\end{figure}

\section{Experiments}

\subsection{Evaluation Metrics}
\label{:subsection:metrics}

\begin{figure}[t]
\begin{center}
\includegraphics[width=.8\textwidth]{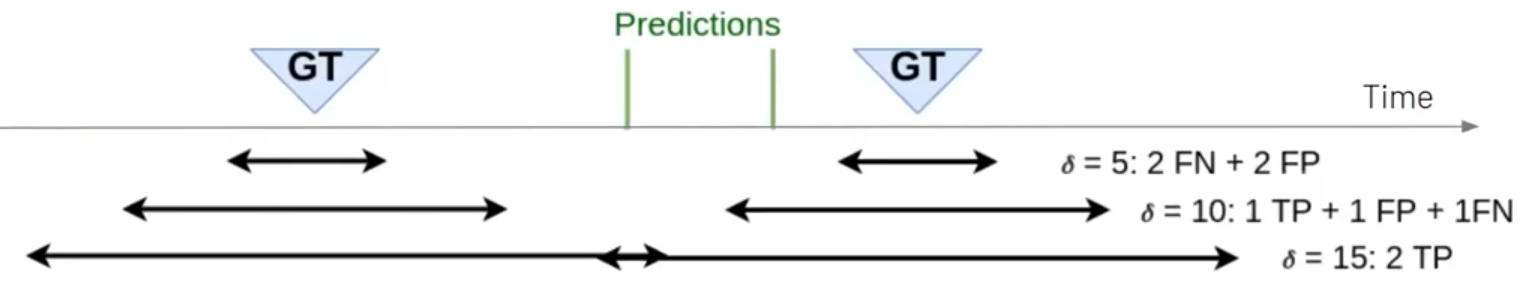}
\end{center}
\caption{A prediction as a True Positive (TP) if it lands within a tolerance of $\delta$ around the ground-truth event. For each class, the AP is calculated by averaging the performance of the model over $\delta$s varying from 5s to 10s as explained in the section \ref{:subsection:metrics}.}
    \label{fig:mAP_calc}
\end{figure}

\begin{figure}[t]
\begin{center}
\includegraphics[width=\textwidth]{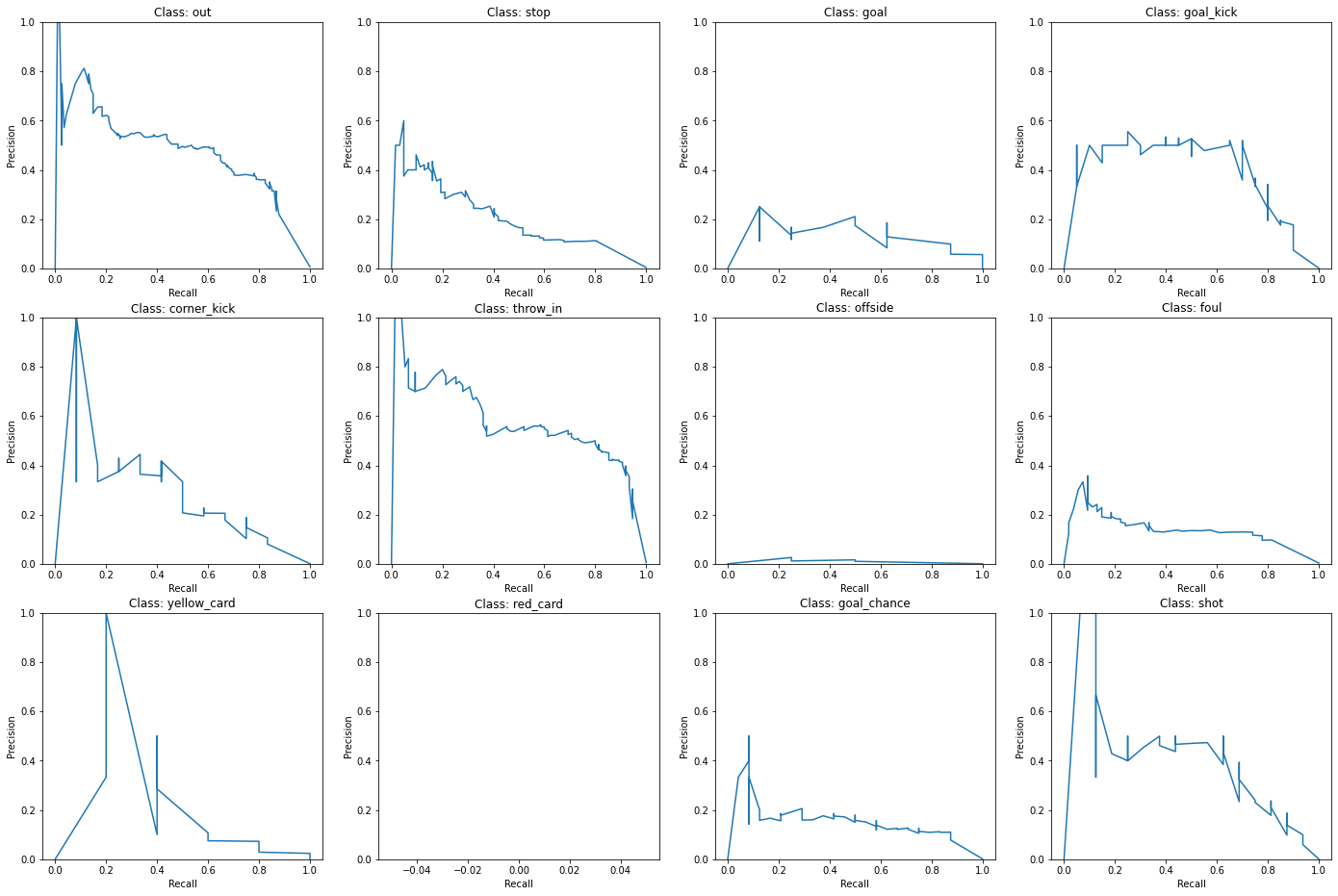}
\end{center}
\caption{Precision-Recall curve computed for each class at a $\delta=30seconds$. Th results are for the test dataset for the best model configuration of 'NetVLAD++', Window size $=$ 10 seconds, and the number of frames per second  $=$ two.}
\label{fig:PR_curves}
\end{figure}

The evaluation follows the mAP metric used in recent papers \cite{soccern_net, soccern_net_v2, calf_loss}. For each class, we mark a prediction as a True Positive (TP) if it lands within a tolerance of $\delta$ around the ground-truth event as can be seen in Fig \ref{fig:mAP_calc}. For each tolerance and class, we then threshold the predictions based on their confidence score and plot a precision-recall (P-R) curve by varying the confidence threshold from $[0,1]$ at 200 points. After obtaining the P-R curve, we modify each point by its right-most max value and then compute the area under this curve by an 11-point approximation $[0, 0.1, 0.2,...,1]$. This area is called the Average Precision of a class at that $\delta$ and can be observed in Fig. \ref{fig:PR_curves}. We then repeat this procedure for a class by varying $\delta$ from 5 seconds to 60 seconds and plot an AP vs $\delta$ curve. Once all APs are computed, we approximate the area under this curve with the trapezoidal formula, and this gives us the average-AP, which can also be called AP for that class. We repeat this procedure for each class and average all the APs to get the mean Average Precision (mAP) of our model.

\subsection{Network Architecture}

The main architecture can be seen in Fig. \ref{fig:summary} and consists of a graph neural network module that is used to extract an embedding from each graph (frame) before it is passed to the event detection model. The graph neural network consists of two Graph Convolutional Layers (GCN) from \cite{kipf2017semi} with hidden dimensions of 64 and 64, followed by a linear layer of size 32. The sizes were chosen keeping in mind the dimensions of the input features to the graph whose dimensions are of size 5 (2 for coordinates and 3 for 1-hot vector). Both the GCN layers are also followed by a batch norm layer and a ReLU non-linearity. The graph embedding is generated by aggregating the feature vector from all the nodes and taking an average. This is referred to as a readout operation in graph convolutional networks.

\begin{figure}[htp]
\centering
\includegraphics[width=.32\textwidth]{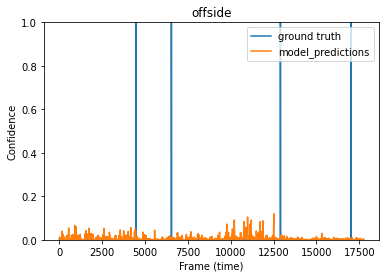}
\includegraphics[width=.32\textwidth]{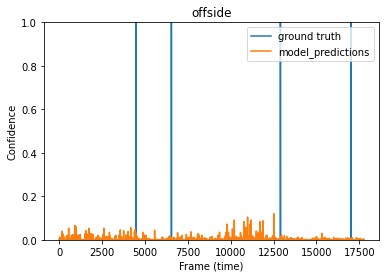}
\includegraphics[width=.32\textwidth]{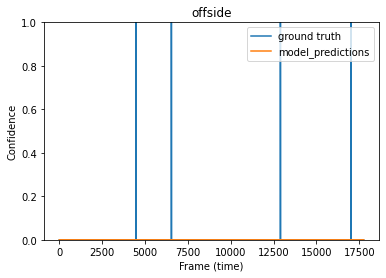}
\includegraphics[width=.32\textwidth]{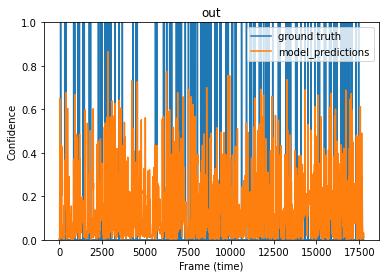}
\includegraphics[width=.32\textwidth]{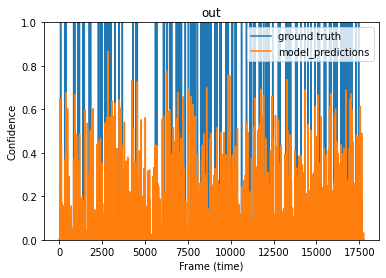}
\includegraphics[width=.32\textwidth]{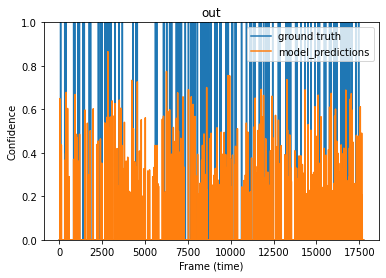}
\includegraphics[width=.32\textwidth]{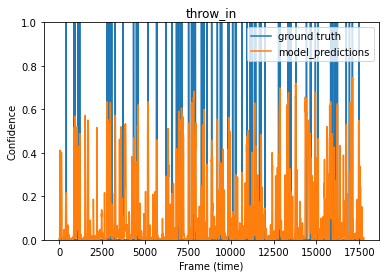}
\includegraphics[width=.32\textwidth]{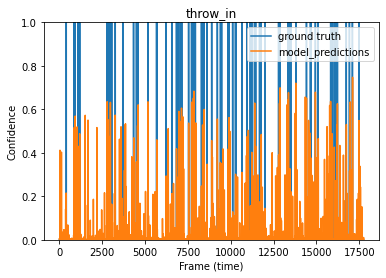}
\includegraphics[width=.32\textwidth]{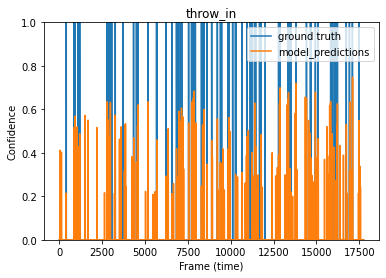}
\includegraphics[width=.32\textwidth]{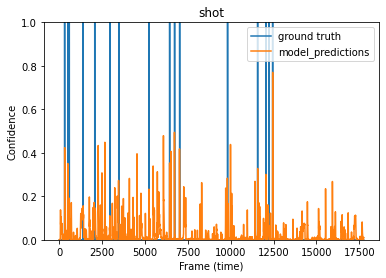}
\includegraphics[width=.32\textwidth]{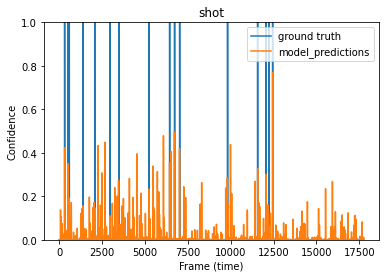}
\includegraphics[width=.32\textwidth]{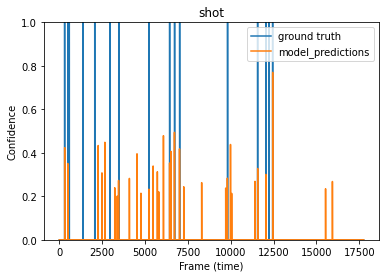}
\caption{All the plots on the first column show the raw predictions outputted per frame using our best model configuration. The plots in the second column show a non-maximum suppression of a 30 seconds window and zero confidence threshold applied to these raw predictions. The third column shows a non-maximum suppression of a 30 seconds window and 0.2 confidence threshold applied to the raw predictions.}
\label{fig:multiple_figures_2}
\end{figure}

\begin{figure}[htp]
\centering
\includegraphics[width=.32\textwidth]{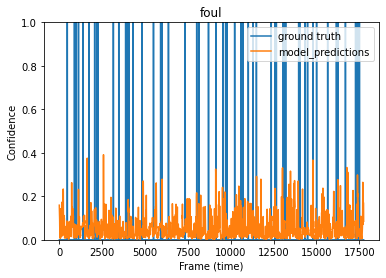}
\includegraphics[width=.32\textwidth]{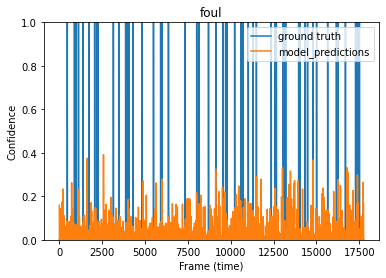}
\includegraphics[width=.32\textwidth]{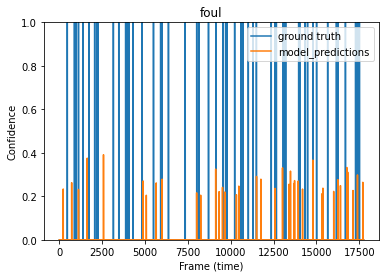}
\includegraphics[width=.32\textwidth]{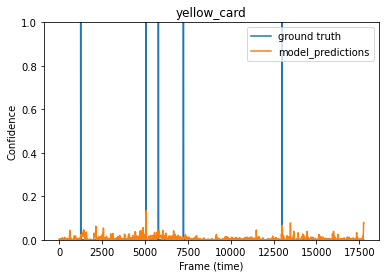}
\includegraphics[width=.32\textwidth]{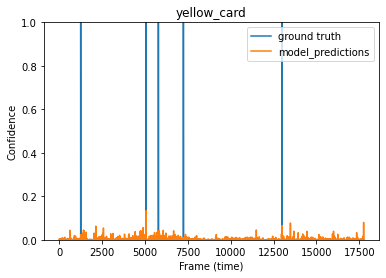}
\includegraphics[width=.32\textwidth]{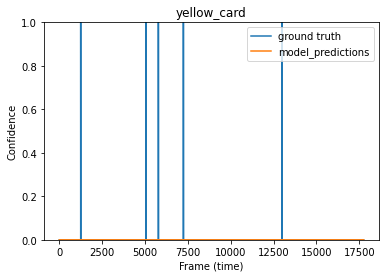}
\includegraphics[width=.32\textwidth]{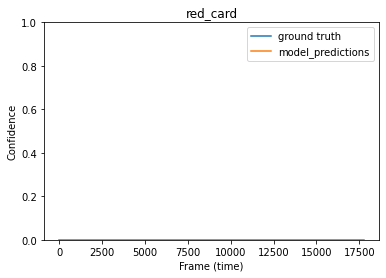}
\includegraphics[width=.32\textwidth]{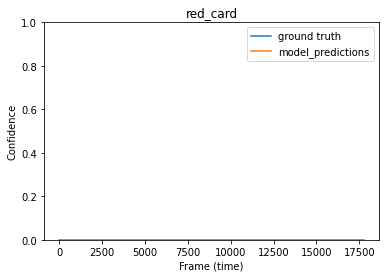}
\includegraphics[width=.32\textwidth]{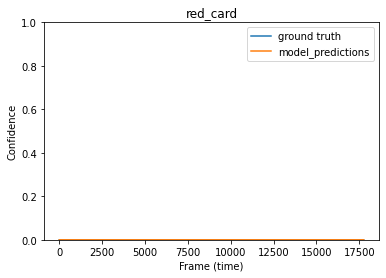}
\caption{All the plots on the first column show the raw predictions outputted per frame using our best model configuration. The plots in the second column show a non-maximum suppression of a 30 seconds window and zero confidence threshold applied to these raw predictions. The third column shows a non-maximum suppression of a 30 seconds window and 0.2 confidence threshold applied to the raw predictions.}
\label{fig:multiple_figures_3}
\end{figure}

We follow an action spotting pipeline similar to the one proposed in SoccerNet \cite{soccern_net}, where they try to classify if an action lies within a certain time window in a multi-label setting. During training, we split our videos into chunks of different lengths annotated with all events occurring within that time window. For each frame in this chunk, an embedding is created using the graph embedding model. We aggregate the features extracted from all the frames present within that window and use a linear layer with sigmoid activation as the last layer, as is usual in the task of multi-label classification. For testing, we sample the frames of the video with the same window size and a stride of one and pass this as input to the model to obtain the raw event predictions for each frame. We then apply a confidence threshold and non-maximum suppression on the output predictions for each class to get the final action spotting results. The results comparing the raw predictions of the model and after applying different settings of non-maximum suppressions and confidence thresholds can be found in the appendix section in Figs. \ref{fig:multiple_figures_1}, \ref{fig:multiple_figures_2} and \ref{fig:multiple_figures_3}. The results for different pooling layers and window sizes can  be found in \ref{tab:results}.

\subsection{Training Setup}\label{subsec:training}

The training, test and validation sets consist of five, two and two matches respectively.

For our temporally aware NetVLAD pooling layers as described in Section \ref{subsection:netvlad++}, we set $K_a = K_b = K/2$ and set $T_a = T_b = T/2$ to consider the same amount of temporal context from before and after the actions. We train our models with the Adam \cite{Adam} optimizer with default parameters from PyTorch \cite{pytorch}, and a starting learning rate of 1e-3. We use ReduceLROnPlateau scheduler from PyTorch which reduces the learning rate by a factor of 10 if the validation loss does not improve for 10 consecutive epochs. This prevents our model from overfitting. We stop the training once the learning rate falls below 1e-8. A single training converges in about $\sim$ 100 epochs and takes approximately 1 second/epoch on a machine with a single NVIDIA Tesla V100 and a 32GB 12-core CPU. Each experiment takes about 10-15 minutes as 90 percent of the time is spent on data creation and loading. During inference, a single game can be processed to obtain the final event predictions in as little as 2 minutes. Note that this time only accounts for the time it requires for processing the tracking data, and not for getting the object detections and generating tracks from raw images, which are part of the pre-processing pipeline.
\section{Results and Discussion}
Results obtained for the dataset described in~\ref{sec:data} and~\ref{subsec:training} using the mAP metric explained in~\ref{:subsection:metrics} are shown in Table~\ref{tab:results}. The upper half of the table clearly shows an improvement when using more sophisticated pooling methods like NetVLAD or NetRVLAD with respect to simpler ones like max pooling or average pooling. All the best mAP results (both per class and total) are obtained with the best pooling methods. It also shows how, in general, the temporally-aware pooling techniques produce better results than the regular ones.

NetVLAD++ shows the best results overall. The bottom half of the table shows an analysis of the effect of using different window sizes for the NetVLAD++ pooling method. Different window sizes produce better results for different classes, but taking into account the duration of the detected events and the contextual information needed for detecting them, 10 seconds looks like a reasonable value and it produces the best results overall.

The model's performance was good on the classes that had a good number of annotations available. However, the events that the model was not able to detect correctly were mostly the ones with very few annotations (less than 50 for some), which is expected since deep learning models require more data for learning good representations.

\begin{table*}[ht]
    \caption{\textbf{Results for different pooling methods.}
    Mean Average Precision (mAP) results for the test dataset using all the different methods trained. Window size is 10 seconds, and the number of frames per second is two. 
    }
    \centering
    \setlength{\tabcolsep}{2pt}
    \resizebox{\linewidth}{!}{
    \begin{tabular}{l||c||c|c|c|c|c|c|c|c|c|c|c|c}
     &  \begin{turn}{90}\textbf{Total} \end{turn} &  \begin{turn}{90} Out \end{turn}   &  \begin{turn}{90} Stop \end{turn}  & \begin{turn}{90} Goal \end{turn} & \begin{turn}{90}Goal kick\end{turn} & \begin{turn}{90} Corner kick \end{turn} & \begin{turn}{90} Throw in \end{turn} & \begin{turn}{90} Offside \end{turn} & \begin{turn}{90} Foul \end{turn} & \begin{turn}{90} Yellow Card \end{turn} & \begin{turn}{90} Red card \end{turn} & \begin{turn}{90} Goal chance \end{turn} & \begin{turn}{90} Shot \end{turn} \\ 

       \midrule \midrule

AVG & 15.2 & 38.6 & 15.3 & 3.4 & 7.0 & 10.5 & 52.5 & 0.1 & 19.3 & 1.1 & 0.0 & 20.7 & 12.8 \\ \midrule
AVG++ & 16.7 & 40.3 & 18.2 & 06.6 & 8.0 & 17.8 & 54.8 & 0.8 & 17.1 & 1.1 & 0.0 & 22.9 & 12.5 \\ \midrule
MAX & 16.4 & 39.5 & 16.0 & 5.8 & 11.0 & 16.1 & 54.6 & 0.6 & 16.1 & 1.0 & 0.0 & 25.0 & 11.8 \\ \midrule
MAX++ & 16.3 & 41.4 & 18.8 & 5.3 & 8.5 & 14.9 & 53.2 & 1.1 & 15.8 & 1.9 & 0.0 & 21.5 & 13.0 \\ \midrule
NetRVLAD & 25.0 & 42.7 & 24.9 & 15.9 & 35.7 & \textbf{38.4} & 55.4 & 1.4 & 24.9 & 23.4 & 0.0 & 20.2 & 17.6 \\ \midrule
NetRVLAD++ & 30.7 & 51.6 & \textbf{32.1} & \textbf{34.5} & 40.8 & 38.1 & 59.5 & 1.0 & 24.8 & 25.1 & 0.0 & 18.9 & 41.6 \\ \midrule
NetVLAD & 27.4 & 45.4 & 25.3 & 8.5 & 57.4 & 36.8 & 56.2 & 0.9 & \textbf{27.3} & 3.5 & 0.0 & \textbf{32.6} & 35.8 \\ \midrule
NetVLAD++ & \textbf{31.3} & \textbf{55.5} & 29.7 & 17.3 & \textbf{50.3} & 35.7 & \textbf{59.7} & \textbf{1.6} & 21.7 & \textbf{36.4} & 0.0 & 20.5 & \textbf{47.0} \\ 
\midrule
\midrule
\textbf{Window Size} & \multicolumn{13}{c}{\textbf{Temporal context}} \\ \midrule \midrule

~~~$\mathbf{T=5s}$ & 26.7 & 51.8 & 24.3 & \textbf{23.5} & \textbf{58.2} & 32.4 & 56.8 & 0.9 & 17.4 & 5.4 & 0.0 & \textbf{23.8} & 25.9 \\ \midrule
~~~$\mathbf{T=10s}$ & \textbf{31.3} & 55.5 & 29.7 & 17.3 & 50.3 & \textbf{35.7} & 59.7 & 1.6 & 21.7 & \textbf{36.4} & 0.0 & 20.5 & \textbf{47.0} \\ \midrule
~~~$\mathbf{T=15s}$ & 28.9 & 54.1 & \textbf{37.6} & 13.8 & 45.9 & 25.8 & \textbf{60.1} & 3.1 & \textbf{30.6} & 18.7 & 0.0 & 19.9 & 37.0 \\ \midrule
~~~$\mathbf{T=20s}$ & 28.7 & \textbf{58.8} & 33.1 & 18.1 & 46.5 & \textbf{35.7} & 54.5 & \textbf{5.8} & 22.1 & 6.3 & 0.0 & 21.8 & 42.2 \\ \midrule
~~~$\mathbf{T=25s}$ & 22.4 & 52.6 & 31.7 & 11.7 & 26.5 & 29.5 & 53.8 & 1.4 & 18.6 & 2.7 & 0.0 & 21.8 & 18.4 \\ \midrule
~~~$\mathbf{T=30s}$ & 18.4 & 44.7 & 26.7 & 7.2 & 19.5 & 23.4 & 50.6 & 3.3 & 16.1 & 3.3 & 0.0 & 17.0 & 9.5 \\ \midrule
~~~$\mathbf{T=35s}$ & 18.4 & 40.8 & 27.7 & 6.3 & 21.9 & 27.4 & 45.0 & 3.2 & 21.4 & 2.4 & 0.0 & 16.4 & 8.9 \\ \midrule
~~~$\mathbf{T=40s}$ & 18.9 & 40.7 & 25.1 & 4.4 & 31.8 & 28.6 & 47.2 & 5.3 & 15.0 & 3.0 & 0.0 & 18.0 & 8.1 \\ \midrule
~~~$\mathbf{T=45s}$ & 15.9 & 37.7 & 21.4 & 7.1 & 11.9 & 23.0 & 39.4 & 1.6 & 16.5 & 3.5 & 0.0 & 16.8 & 12.0 \\ \midrule
~~~$\mathbf{T=50s}$ & 18.6 & 42.3 & 26.9 & 6.5 & 22.3 & 23.7 & 44.5 & 4.6 & 17.3 & 4.1 & 0.0 & 19.2 & 12.2 \\ \midrule
~~~$\mathbf{T=55s}$ & 16.3 & 36.3 & 24.9 & 5.5 & 18.8 & 22.2 & 40.7 & 2.3 & 12.6 & 5.8 & 0.0 & 18.1 & 8.9 \\ \midrule
~~~$\mathbf{T=60s}$ & 16.1 & 35.6 & 23.3 & 5.3 & 11.8 & 24.3 & 41.4 & 2.7 & 13.6 & 13.4 & 0.0 & 13.9 & 7.6 \\ \bottomrule

\end{tabular}}
\label{tab:results}
\end{table*}

\section{Conclusions}

We presented a methodology for detecting events based on processing tracking data that is generated from the players and discussed how graph neural networks can overcome the problems faced by other machine learning models when processing this type of data because it is unclear on how to order players in a sequence and how to handle missing objects of interest when constructing a feature vector summarizing a frame. We show how to model the players and the ball in each frame of the video sequence as a graph, and discuss how the performance of pooling layers in event detection models can be improved by considering the context before and after the action separately. We were able to get good results for few classes despite having just a few annotations for each class compared to other recent works which are trained on datasets with thousands of annotations per event. For future work, we would like to explore self-supervised techniques for pre-training our graphs before training them on event detection tasks. One of the tasks we have in mind is predicting the future motion of teams given the previous positions of its players over a window.

\section*{Acknowledgements}

There are many people I must thank for the development of this thesis and the wonderful year that I have spent studying and working in Barcelona. Firstly, I would like to thank my supervisor, Dr. Francesc Moreno, without whose trust and support this project would not have been possible. I am grateful to him for allowing me to intern at Institut de Robòtica i Informàtica Industrial as well as giving me the chance to continue working under his supervision at Kognia Sports Intelligence, where this thesis has been developed. I feel extremely lucky to have Dr. Antonio Rubio Romano as my co-supervisor, under whose mentorship I have been able to learn and grow immensely in the past year. Antonio is one of the most insightful people I know as he forces you to increase your understanding of anything by questioning the very fundamentals of your ideas. I want to thank him for all the hours he has helped me in writing, debugging my code, and helping me become a much better programmer. I would also like to thank all the people at Kognia for making my previous year full of learning and hard work, especially Dr. Luis Ferraz who has been an amazing team leader. I would also like to thank all the professors who taught in this Master's program, as they helped me build my ideas in machine learning and computer vision from the very fundamentals. I am extremely grateful to them for all the effort they put in teaching us as well as solving our doubts. A big thank you to my family for believing in me much more than I do and for making so many sacrifices to ensure I receive the best of opportunities in life. It is, in large part, my determination to vindicate their leap of faith and make them proud that drives my ambitions.

\clearpage

\appendices
\section{Confidence scores}
\begin{figure}[htp]
\centering
\includegraphics[width=.32\textwidth]{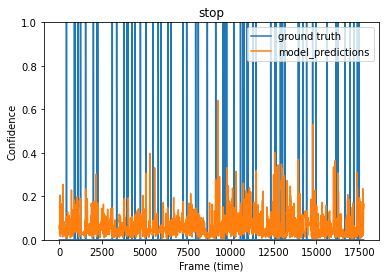}
\includegraphics[width=.32\textwidth]{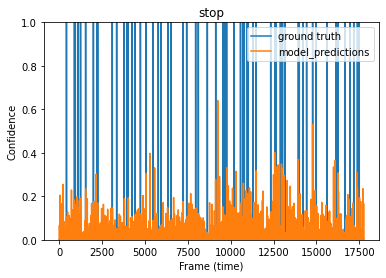}
\includegraphics[width=.32\textwidth]{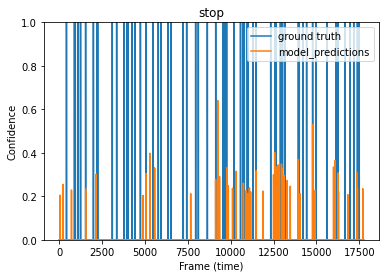}
\includegraphics[width=.32\textwidth]{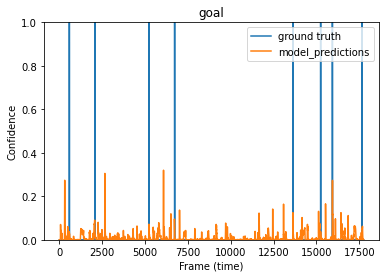}
\includegraphics[width=.32\textwidth]{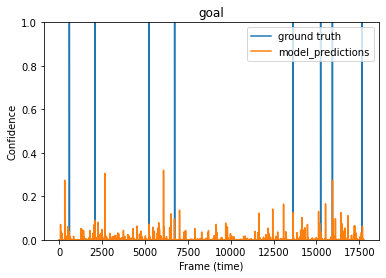}
\includegraphics[width=.32\textwidth]{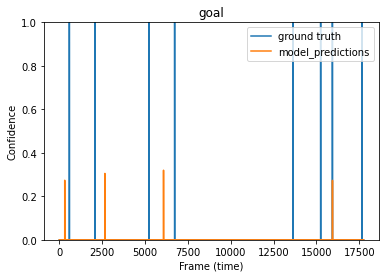}
\includegraphics[width=.32\textwidth]{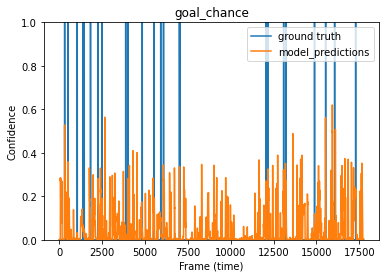}
\includegraphics[width=.32\textwidth]{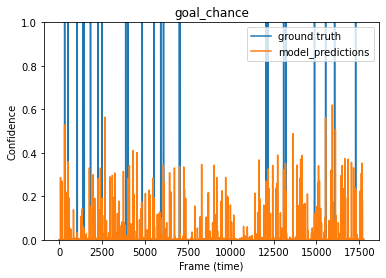}
\includegraphics[width=.32\textwidth]{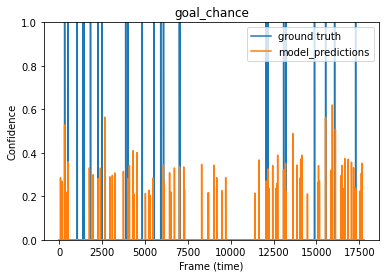}
\includegraphics[width=.32\textwidth]{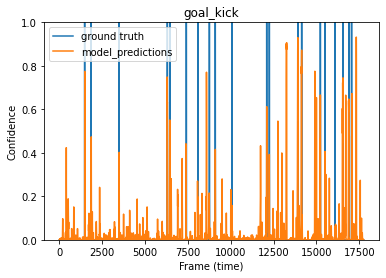}
\includegraphics[width=.32\textwidth]{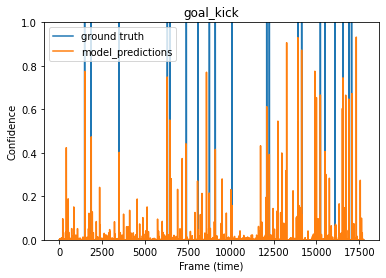}
\includegraphics[width=.32\textwidth]{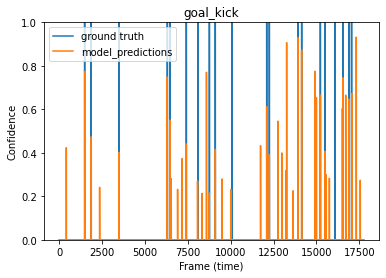}

\caption{All the plots on the first column show the raw predictions outputted per frame using our best model configuration. The plots in the second column show a non-maximum suppression of a 30 seconds window and zero confidence threshold applied to these raw predictions. The third column shows a non-maximum suppression of a 30 seconds window and 0.2 confidence threshold applied to the raw predictions.}
\label{fig:multiple_figures_1}
\end{figure}

\ifCLASSOPTIONcaptionsoff
  \newpage
\fi

\clearpage

\bibliography{ref}{}
\bibliographystyle{IEEEtran}
\end{document}